\documentclass{ifacconf}

\usepackage{graphicx}      
\usepackage{natbib}        
\usepackage{xcolor}
 \usepackage{color,soul}
\usepackage{amsmath} 
\usepackage{amssymb}
\usepackage{booktabs}
\begin{document}
\begin{frontmatter}

\title{Geometric Graph Neural Network Modeling of Human Interactions in Crowded Environments \thanksref{footnoteinfo}} 

\thanks{This work was supported by the Diversity in Robotics and AI PhD Fellowships by Microsoft and Amazon Lab 126 from the Maryland Robotics Center. }


\author[First]{Sara Honarvar} 
\author[First]{Yancy Diaz-Mercado} 

\address[First]{Department of Mechanical Engineering, University of Maryland, College Park, MD 20742 USA (e-mail: honarvar@ umd.edu, yancy@ umd.edu)}
\thanks[footnoteinfo]{© 2024 the authors. This work has been accepted to IFAC for publication under a Creative Commons Licence CC-BY-NC-ND.}
\begin{abstract}                
Modeling human trajectories in crowded environments is challenging due to the complex nature of pedestrian behavior and interactions. This paper proposes a geometric graph neural network (GNN) architecture that integrates domain knowledge from psychological studies to model pedestrian interactions and predict future trajectories. Unlike prior studies using complete graphs, we define interaction neighborhoods using pedestrians' field of view, motion direction, and distance-based kernel functions to construct graph representations of crowds. Evaluations across multiple datasets demonstrate improved prediction accuracy through reduced average and final displacement error metrics. Our findings underscore the importance of integrating domain knowledge with data-driven approaches for effective modeling of human interactions in crowds.
\end{abstract}


\end{frontmatter}

\section{Introduction}



Over the last decade, predicting how humans move and interact has become an active field of research, driven by applications in autonomous driving, surveillance, and the creation of socially aware robots (see \cite{rudenko2020human, mavrogiannis2023core} and references therein). The challenge lies in anticipating human motion due to its inherently complex and stochastic nature, influenced by internal and external stimuli, social dynamics, and psychological decision-making factors as described in \cite{wirth2022neighborhood}. This challenge is magnified in crowded spaces, where the interplay of these factors becomes even more intricate (see \cite{korbmacher2022review} for a review).

Despite this complexity, humans demonstrate an impressive ability to navigate crowds. As we move, we instinctively stay close to our group while maintaining a social distance from others. Our behavior adapts based on the type of interaction at play. Understanding these implicit interactions is crucial for autonomous systems, enabling them to predict and adapt to the unpredictable behaviors of pedestrians in the absence of explicit communication. 

Various modeling-based and learning-based algorithms have been proposed to understand human interactions when navigating through crowds, as reviewed in \cite{korbmacher2022review}. Model-based approaches, such as the social force model by \cite{helbing1995social} and velocity-based methods like the reciprocal n-body collision avoidance technique introduced by \cite{van2011reciprocal}, aim to develop intuitive strategies for explaining human interactions based on first principles and social factors. In contrast, learning-based methods leverage deep learning algorithms, including long short-term memory (LSTM) networks \cite{alahi2016social}, convolutional neural networks (CNNs), generative adversarial networks (GANs) \cite{gupta2018social}, and graph neural networks (GNNs) \cite{mohamed2020social}, to predict human motions directly from data.

Both model-based and learning-based approaches have their advantages and disadvantages. Model-based algorithms typically have fewer parameters, making them computationally efficient and enabling careful reasoning about human intentions. However, they may overlook complex aspects of human behavior. 
Our proposed model takes the form of a geometric graph neural network, a structure inspired by the Social Spatio-Temporal Graph Neural Network (STGCNN) architecture proposed by \cite{mohamed2020social}. However, we go beyond conventional GNN formulations by integrating geometric kernel functions and a refined neighborhood of interaction definition, drawing insights from psychological studies on human crowd behaviors (\cite{rio2018local, wirth2022neighborhood}). Our hybrid approach aims to combine the strengths of both model-based and learning-based methods by incorporating geometric kernel functions and refined interaction neighborhood definitions derived from psychological studies.  These enhancements contribute to a more interpretable and context-aware weighted adjacency matrix, enabling a comprehensive framework for modeling pedestrian interactions.
\section{Related work}
In recent years, deep learning approaches for predicting human motion have gained significant attention. These methods typically assume that pedestrian trajectories follow a bivariate Gaussian distribution and aim to predict trajectory distributions based on observed frames.

Social-LSTM proposed by \cite{alahi2016social} represents one of the early methods using a recurrent neural network to predict individual pedestrian motion. To incorporate human interactions, it introduces a social pooling layer that aggregates recurrent outputs and concatenates hidden states of nearby agents. Extending this idea, Social-GAN by \cite{gupta2018social} employs generative networks and a loss function encouraging multi-modality of pedestrian trajectory distributions, as agents' trajectories may have multiple possible futures. However, it assigns uniform weights to all surrounding pedestrians, failing to capture the importance of close neighbors.

Given the advancements of GNNs introduced in \cite{kipf2016semi} for explaining and modeling interactions between different entities, recent works have integrated graph representation GNNs to capture the interaction mechanism between humans, departing from traditional pooling methods (see \cite{korbmacher2022review} for an extensive review). 

GNN is a powerful tool for learning representations of network systems, but they rely heavily on the given network structure which may not accurately reflect the true interactions. Most approaches construct an initial graph based on node feature similarities, then refine it using metric (e.g., neural (e.g., attention mechanism such as the works by \cite{huang2019stgat} and \cite{kosaraju2019social}), kernel functions as in \cite{mohamed2020social}), or direct methods (e.g., optimization methods enumerated in \cite{zhu2021survey}).

For instance, STGAT proposed by \cite{huang2019stgat} and social-BiGAT by \cite{kosaraju2019social} treat each agent as a node in a graph, and use Graph Attention Networks (GAT) to describe their interaction by weighing the importance of each pedestrian's contribution. Although attention mechanism is beneficial in terms of modeling the importance of each neighboring pedestrian, they exhibit drawbacks such as a lack of explainability and a substantial increase in training time and learnable parameters. In response to these limitations, alternative approaches like social STGCNN (\cite{mohamed2020social}) have emerged. This method proposes a graph representation for modeling crowd interaction based on spatio-temporal information, opting for a kernel function instead of an attention mechanism to weigh the importance of agents before constructing a graph representation of pedestrians. This strategic choice not only overcomes the shortcomings associated with attention mechanisms but also introduces a more interpretable and computationally efficient approach to assessing the importance of individual agents within a crowd.

Our work builds upon the kernel-based GNN framework proposed in Social STGCNN by \cite{mohamed2020social}, but explores the impact of various geometric kernel functions to identify the neighborhood of interaction in crowds, determining which neighbors influence each pedestrian's behavior and to what extent. These proposed kernel functions draw inspiration from psychological studies (\cite{rio2018local}) that conducted experiments on human crowds to understand the underlying mechanisms of their interactions. By leveraging this domain knowledge, we aim to enhance the model's ability to reason about pedestrian interactions in a more interpretable and context-aware manner, paving the way for improved prediction and navigation capabilities in crowded environments.


\section{Method}
\subsection{Problem Formulation}
The main objective in this paper is to predict the future trajectory positions of $N$ pedestrians over a time horizon of $T_{\text{pred}}$, given their observed trajectories from $t=1$ to $t=T_{\text{obs}}$. This task is particularly challenging in crowded environments, where pedestrian motion is governed by complex interactions and social dynamics.


\subsubsection{\textbf{Agent Dynamics:}}
Following the work in \cite{mohamed2020social}, we assume pedestrian trajectories follow a bi-variate Gaussian distribution. The parameters $(\mu_t^i,\sigma_t^i,\rho_t^i)$ represent the mean, standard deviation, and correlation coefficient of the Gaussian distribution, respectively, where we parameterize mean $\mu_t^i=(\mu_x,\mu_y)^i_t$, standard deviation $\sigma_t^i=(\sigma_x,\sigma_y)^i_t$, and correlation coefficient $\rho_t^i=(\rho_x,\rho_y)^i_t$, for $i= {1, \dots, N}$. The pedestrian trajectories are given by $p_t^n=(x^i_t,y^i_t) \sim \mathcal{N}(\mu_t^i,\sigma_t^i,\rho_t^i)$, where $(x^i_t,y^i_t)$ are random variables capturing the probability distribution of pedestrian $i$'s position at time $t$ in the 2D space. The model is trained by minimizing the negative log-likelihood loss function:
\begin{equation}
J^i(W) = - \sum_{t=1+T_{\text{obs}}}^{T_{\text{pred}}} \log(\mathbb{P}(p_t^n|\hat{\mu}_t^i,\hat{\sigma}_t^i,\hat{\rho}_t^i)),
\end{equation}
Here, $\hat{\mu}_t^i,\hat{\sigma}_t^i,\hat{\rho}_t^i$ are  the estimated parameters of the predicted trajectory distribution, $\hat{p}_t^i$.
\subsubsection{\textbf{Graph Representation of Crowd Interaction:}}

We employ a graph representation rooted in algebraic graph theory to model pedestrian interactions within crowd \cite{mesbahi2010graph}. Consider $G_t=(\mathcal{V}_t,E_t)$ as a spatial graph at time $t$, comprising $N$ nodes representing the relative positions of pedestrians in the scene at time $t$. The node set is $\mathcal{V}_t=\{\nu_t^1,\dots,\nu_t^N\}$, and edge set $E=\{e_t^{ij}=(\nu_t^i,\nu_t^j)\} \subseteq \mathcal{V}_t \times \mathcal{V}_t$ captures the interactions among agents. The adjacency matrix is defined as $A_t=[a_t^{ij}]$, determining the neighborhood of interaction, where $a_t^{ij}=w_t^{ij}$ if $\nu_t^i$ and $\nu_t^j$ are connected (agent $i$ can receive information from agent $j$); otherwise, $a_{ij}=0$. The interaction neighborhood will be defined in more details later. The weights $w_t^{ij}$ quantify how strongly agents influence each other's movements and are determined through geometric kernel functions inspired by psychological studies on human crowd behaviors. The graph Laplacian is defined as $L_t = D_t - A_t$, where $D_t$ denotes the diagonal matrices of node degrees, taken as the sum of the weights for each neighbor: $D_t = [d_t^{ij}] = \sum_{j=1}^N a_t^{ij}$.
\begin{figure}[t!]
\begin{center}
\includegraphics[width=0.28\textwidth]{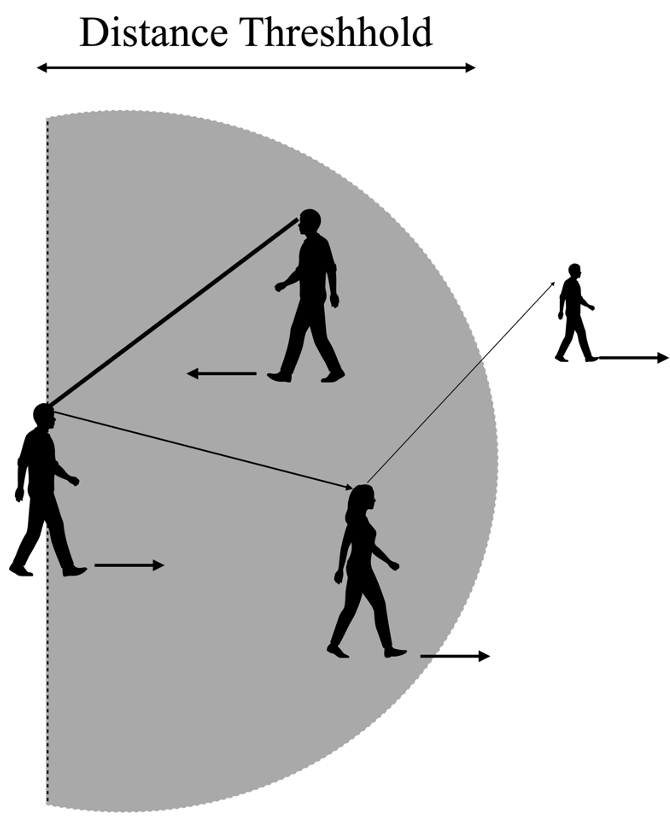}    
    \caption{A visual representation of the neighborhood of interaction for a pedestrian (shown in black), incorporating a 180-degree field of view and a 5-meter distance threshold, selected based on empirical observations from psychological studies as described in \cite{rio2018local}. Pedestrians within the shaded region, which extends up to 90 degrees on either side of the walking direction, are considered part of the neighborhood. The thickness of the connecting lines decreases with increasing distance, indicating that closer neighbors have a stronger influence on the pedestrian's motion (corresponding to Graph 2 neighborhood definition with distance-based weighting according to our kernel definitions)}
\label{fig:graph2}
\end{center}
\end{figure}

\subsection{Topology of Interaction}

Diverging from the approach of social-STGCNN \cite{mohamed2020social}, which assumes a complete graph to establish the neighborhood set for each agent and weighs the importance of neighboring agents based solely on their distance, this work takes a more nuanced perspective. Our goal is to experiment with different neighborhood sets 
inspired by psychological literature on human interactions in crowds, as described in \cite{rio2018local}, aiming to enhance model performance in predicting future pedestrian motion. This exploration seeks to identify the crucial factors in accurately modeling human interactions across various datasets.

\emph{\textbf{Neighborhood of Interaction:}}

According to \cite{grasso1998eye}, people have a $180^\circ$ horizontal field of view and face toward their walking direction ($\Delta p_t^i = p_{t+1}^i-p_{t}^i$). Therefore, interactions are expected to be influenced by neighbors within their field of view ($\pm 90^\circ$) who are approaching them. Empirical experiments on human crowds by \cite{rio2018local} reveal that the impact of neighbors diminishes exponentially to zero at a distance of $4–5 m$. The following sections explain how we incorporate these key points in constructing our weighted adjacency matrix.

\begin{enumerate}
    \item \emph{\textbf{Graph 1 (View):}}

To account for the $180 ^\circ $ field of view, we introduce the following kernel-based weight. It stipulates that if the angle between walking direction of pedestrian $i$ and $j$ is smaller than $90 ^\circ $, they mutually influence each other; otherwise, the weight is set to zero.
\begin{align}
    w_{view,t}^{ij} = \begin{cases}
        K_{ij} & \Delta p_t^i \cdot \Delta p_t^j>0 \\
        0 & \text{o.w.}
    \end{cases}.
\end{align}
$K_{ij}$ will be defined shortly.

\item \emph{\textbf{Graph 2 (View, Thresh):}}

In a different scenario, we incorporate a threshold of $\epsilon =5 m$ distance, selected based on empirical observations from psychological studies according to \cite{rio2018local} to ensure that only nearby pedestrians within the field of view influence each other (Fig. ~\ref{fig:graph2}):
\begin{align}
    w_{view, thresh,t}^{ij} = \begin{cases}
        K_{ij} &  \Delta p_t^i \cdot \Delta p_t^j>0, 
        d_t^{ij} <\epsilon\\
        0 & \text{o.w.}
    \end{cases}
    \end{align}
where $d_t^{ij}=\|p_t^i-p_t^j\|$ is the distance between agent $i$ and $j$.

\item \emph{\textbf{Graph 3 (Approach):}}

In this scenario, when the distance between agents decreases, signifying that they are approaching each other and likely to influence each other, we define the kernel-based weight as:
\begin{equation}
   w_{appr,t}^{ij}  = \begin{cases}
        K_{ij} &  d_{t+1}^{ij} > d_t^{ij} \\
        0 & \text{o.w.}
    \end{cases}
\end{equation}

\item \emph{\textbf{Graph 4 (View, Approach):}}
We also consider a case that encompasses both of the above scenarios:
\begin{equation}
    w_{view, appr,t}^{ij} = \begin{cases}
        K_{ij} &  \Delta p_t^i \cdot \Delta p_t^j>0, 
        d_{t+1}^{ij} > d_t^{ij} \\
        0 & \text{o.w.}
    \end{cases}
\end{equation}
\end{enumerate}
It is worth noting that \cite{yang2023trajectory} also considers the field of view and direction of movement. However, they consider each of the factors in separate graphs, and then use a multi-layer perceptron (MLP) to combine them into a single graph. In this work, we choose to combine the two conditions into a single graph to follow psychological studies and better explain the adjacancy matrix.

\emph{\textbf{Kernel Functions for Assigning Neighbors Importance Through Edge Weights:}}

We employ kernel functions to capture the strength of interaction between pedestrians. We examine the impact of two distinct kernel functions on the model's accuracy in predicting pedestrian trajectories.

\begin{enumerate}
    \item \textbf{Kernel 1 (Inverse of Norm):} 
    Similar to \cite{mohamed2020social}, we use the inverse of the $L_2$ norm as a similarity measure, considering that further away neighbors exert less influence:  $$K_{ij} = \begin{cases}
     \frac{1}{\|p_t^i-p_t^j\|} & \|p_t^i-p_t^j\| \neq 0\\
     0 & \textit{o.w}.
    \end{cases}$$
    \item Alternatively, in this case, we use the exponential decay as the kernel function:
    $$K_{ij} = \begin{cases}
    \exp (-\|p_t^i-p_t^j\|) & \|p_t^i-p_t^j\| \neq 0\\
    0 & \textit{o.w}.
    \end{cases}$$
    to reflect the fact that the weight decays as agents move away from each other, according to \cite{rio2018local}.
\end{enumerate}

\begin{figure*}[t!]
\begin{center}
\includegraphics[width=1\textwidth]{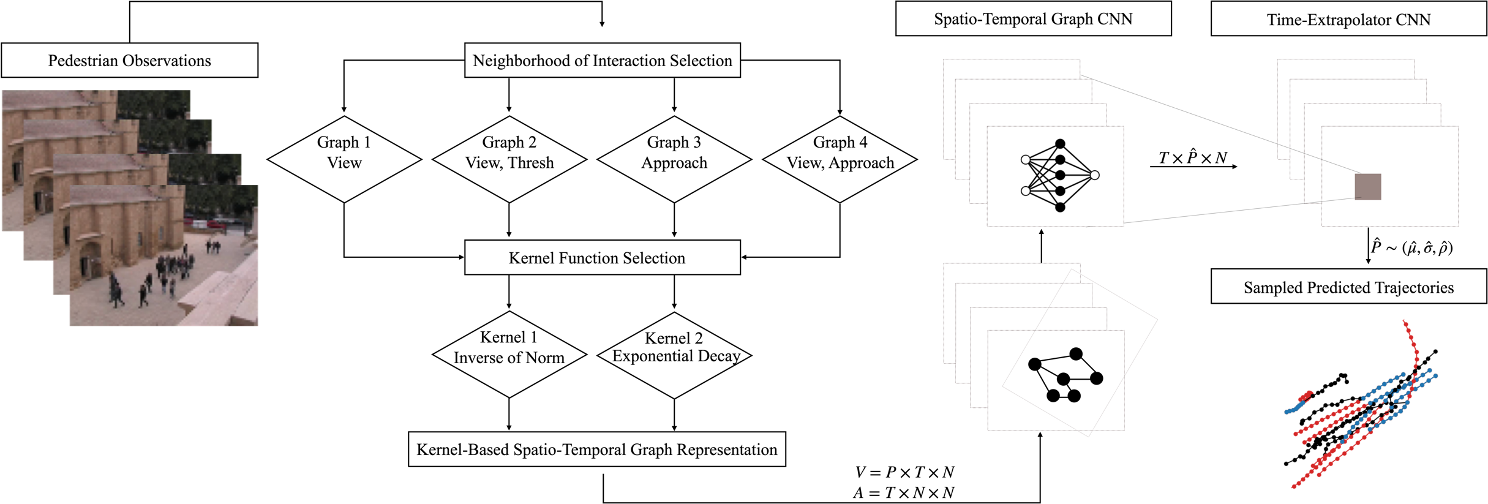}    
    \caption{Illustration of our geometric GNN model. From observed pedestrian trajectories over $T$ frames, we construct a kernel-based spatio-temporal graph by defining neighborhood relationships and weighting edges using kernel functions. This graph is processed by a spatio-temporal graph CNN, similar to the architecture proposed by \cite{mohamed2020social}, to generate a spatio-temporal embedding. A temporal CNN then predicts the distribution of future trajectories. $V$ represents the set of nodes, $P=2$ is the dimension of pedestrian positions, and $\hat{P}$ denotes the dimensions of the embedding from the predicted Gaussian distribution. Our approach integrates domain knowledge from psychological studies, resulting in a more context-aware and interpretable adjacency matrix.}
\label{fig:model}
\end{center}
\end{figure*}
\subsection{\textbf{Model Architecture:}}
Our model leverages the same network architecture as introduced in \cite{mohamed2020social}, consisting of spatial and temporal components. The key difference from previous work is in the graph representation, which incorporates geometric kernels and neighborhood definitions inspired by psychological studies on crowd dynamics, as explained earlier. Fig.~ \ref{fig:model} provides a visual overview of the network architecture adopted in this paper.
\begin{enumerate}
    \item \textbf{Kernel-based Spatio-Temporal Graph Representation:}
    The first step in our network is to construct the graph representation of pedestrian interactions. We utilize the proposed neighborhood of interactions and kernel functions one by one and report the modeling accuracy for each of them.

\item \textbf{Spatio-Temporal Graph CNN:}

In the second step, to extract meaningful features from the graph representation, we employ spatio-temporal graph convolutions, as described by \cite{yan2018spatial}. This approach is particularly suitable for our problem as it effectively captures both spatial and temporal dependencies in pedestrian trajectories. The colvolution operation compute node embeddings by aggregating information from neighboring nodes in a recursive fashion as follows:
\begin{equation}\label{convolution}
    f(V^{(l)}, A) = \sigma\left(\mathcal{L}_t V^{(l)} W^{(l)}\right),
\end{equation}
where $V^{(l)}$ is the stack of all node values at time $t$ and layer $l$, and $W^{(l)}$ is the matrix of trainable parameters at layer $l$. $\mathcal{L}_t = D_t^{-1/2}L_tD_t^{-1/2}$ is the normalized graph Laplacian matrix at time $t$, as explained in \cite{kipf2016semi}. Finally, $\sigma$ is the activation function.

\item \textbf{Time-Extrapolator CNN:}

Taking the features from the previous step as input, this network applies temporal convolutions to predict probability distributions of future trajectories for all pedestrians within the time interval from $t=T_{\text{obs}+1}$ to $t=T_{\text{pred}}$.
\end{enumerate}

\subsection{Datasets and evaluation metrics}
To train and assess the performance of our model, we utilize two well-established pedestrian datasets, in line with established methodologies in prior research. These datasets include: (1) ETH Dataset, collected by \cite{pellegrini2009you}, with two scenes: ETH and Hotel, and (2) UCY Dataset, given by \cite{lerner2007crowds}:
with 3 scenarios: ZARA-01, ZARA-02, and UNIV. In total, our model is evaluated across five distinct datasets, encompassing diverse nonlinear trajectories from 1536 pedestrians within real-world crowded environments. To ensure consistency in data representation across various datasets, we adopt the preprocessing approach outlined in Social-GAN by \cite{gupta2018social}. To facilitate fair comparisons, we employ a leave-one-out training methodology, as suggested \cite{alahi2016social}. This involves training and validating on four datasets and subsequently testing on the fifth. This procedure is consistently applied across the \textbf{four} interaction neighborhood scenarios and the \textbf{two} proposed kernel functions as well as other baseline methods for best evaluation.

For the quantitative assessment of our approach, we employ two widely recognized metrics \cite{alahi2016social}:

\ul{Average Displacement Error (ADE)} computes the average L2 distance between the ground truth and our model's predictions over all predicted time steps. It serves as a measure of trajectory accuracy along the path.

\ul{Final Displacement Error (FDE)} quantifies the distance between the predicted final position and the ground truth final position at the end of the prediction period $T_{\text{pred}}$.

Given our assumption of Gaussian distribution for pedestrian trajectories, the estimated trajectory is a distribution. For fair comparison with previous work, we follow \cite{mohamed2020social}'s evaluation approach. We generate 20 samples from the predicted trajectory distribution and compare the best-performing sample (closest to the ground truth) against the ground truth to calculate ADE and FDE. This evaluation approach has been widely adopted by previous studies, including those by \cite{gupta2018social, huang2019stgat}.


\subsubsection{\textbf{Training setup:}}
Our training setup closely aligns with the architecture and parameters outlined in \cite{mohamed2020social}, ensuring a consistent baseline for evaluation. We employ the Parametric Rectified Linear Unit (PReLU) as our activation function, denoted by $\sigma$. The training batch size is set to 128 and the model is trained for a total of 250 epochs for each dataset. The learning rate is initialized at 0.01 and adjusted to 0.002 after 150 epochs to enhance the model's adaptability over the training duration. Stochastic Gradient Descent (SGD) is employed for optimization, in line with the methodology proposed by \cite{mohamed2020social}. Consistent with the referenced work, our model comprises 1 layer of Spatio-Temporal Graph CNN and 5 layers of Temporal-Extrapolator CNN. This allows evaluating the impact of kernel functions and interaction topologies.
\begin{table*}[ht]
\begin{center}
\caption{ADE/FDE Results for Various Methods across 5 Pedestrian Datasets}
\label{tab:results}
\begin{tabular}{lcccccc}
\toprule
\textbf{Method} & \textbf{ETH} & \textbf{HOTEL} & \textbf{UNIV} & \textbf{ZARA01} & \textbf{ZARA02} & \textbf{Average} \\
\midrule
social-LSTM (\cite{alahi2016social}) & 1.09/2.35 & 0.79/1.76 & 0.67/1.40 & 0.47/1.00 & 0.56/1.17 & 0.72/1.54 \\
social-GAN (\cite{gupta2018social}) & 0.81/1.52 & 0.72/1.61 & 0.60/1.26 & 0.34/0.69 & 0.42/0.84 & 0.58/1.18 \\
Social-BiGAT (\cite{kosaraju2019social}) & 0.69/1.29 & 0.49/1.01 & 0.55/1.32 & \textbf{0.30}/0.62 & 0.36/0.75 & 0.48/1.00 \\
social-STGCNN (\cite{mohamed2020social}) & \textbf{0.64}/1.11 & 0.49/0.85 & \textbf{0.44}/\textbf{0.79} & 0.34/0.53 & 0.30/0.48 & 0.44/0.75 \\
\midrule
\textit{Ours (Kernel 1: Inverse of Norm)} \\
- View (Graph 1- Kernel 1)& \textbf{0.64}/\textbf{1.09} & 0.41/0.63 & \textbf{\textcolor{gray}{0.45}}/0.85 & 0.35/0.59 & 0.30/0.5 & \textbf{0.43}/\textbf{0.73} \\
- View, thresh (Graph 2- Kernel 1) & 0.76/1.48 & 0.40/0.69 & \textbf{\textcolor{gray}{0.45}}/0.85 & 0.33/\textbf{0.52} & 0.29/0.47 & 0.44/0.80 \\
- Approach (Graph 3- Kernel 1)& 0.82/1.29 & 0.39/0.58 & 0.46/0.868 & 0.34/0.53 & 0.30/0.50 & 0.46/0.75 \\
- View, approach (Graph 4- Kernel 1) & 0.75/1.37 & 0.38/0.57 & 0.47/0.88 & 0.34/0.57 & 0.30/0.50 & 0.44/0.77 \\
\midrule
\textit{Ours (Kernel 2: Exponential Decay)} \\
- View (Graph 1- Kernel 2)& 0.80/1.54 & 0.47/0.77 & 0.46/0.84 & 0.37/0.58 & 0.43/0.67 & 0.51/0.88 \\
- View, thresh (Graph 2- Kernel 2)& 0.80/1.40 & 0.47/0.77 & \textbf{\textcolor{gray}{0.45/0.82}} & \textbf{\textcolor{gray}{0.32}}/0.53 & \textbf{0.27}/\textbf{0.45} & 0.47/0.80 \\
- Approach (Graph 3- Kernel 2)& 0.86/1.37 & \textbf{0.34}/\textbf{0.47} & 0.47/0.86 & 0.36/0.56 & 0.28/0.46 & 0.47/0.74 \\
- View, approach (Graph 4- Kernel 2)& 0.76/1.36 & 0.47/0.83 & 0.48/0.88 & 0.33/\textbf{0.52} & 0.30/0.48 & 0.47/0.81 \\
\bottomrule
\end{tabular}
\end{center}
\end{table*}
\begin{figure}
    \centering
    \includegraphics[width=0.40\textwidth]{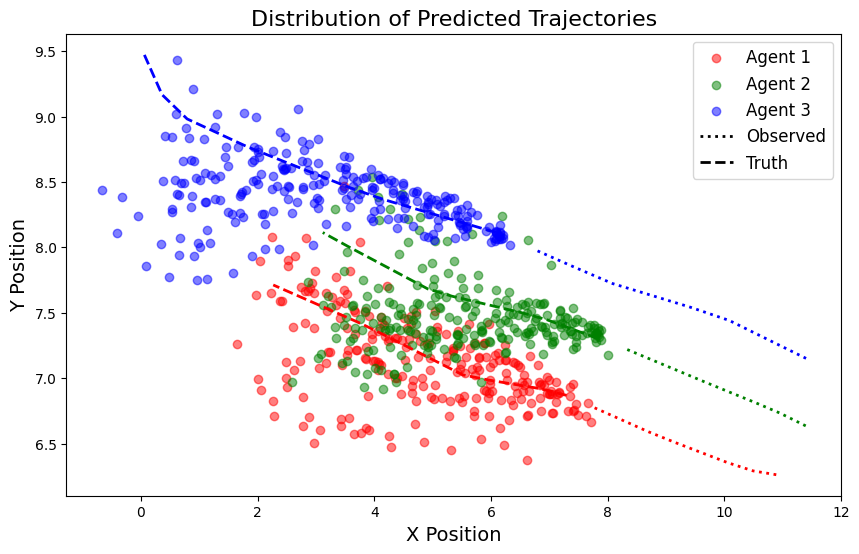}
    \caption{Visualizing the predicted distribution of pedestrian trajectories for a representative scene (UNIV). Sampled trajectories from the predicted Gaussian distribution are depicted as dot points in the scatter plot, while observed and true trajectories are shown as dashed and dotted lines respectively.}
    \label{fig:predicted trajectory}
\end{figure}
\section{Results}
The results presented in Table \ref{tab:results} show that our proposed methods, incorporating different neighborhood definitions and kernel functions, achieve competitive performance compared to previous approaches across various pedestrian datasets. Our model has $7,563$ parameters, with an average inference time of $0.002$ seconds, based on multiple single inference steps.

On the ETH dataset, our method with Graph 1 (using pedestrian's field of view in neighborhood definition) and Kernel 1 (inverse of the norm) achieves the best performance, following the Social-STGCNN ADE of 0.64 and improving the FDE to 1.09.
For the HOTEL dataset, our method with Graph 3 (using approach dynamics in neighborhood definition) and Kernel 2 (exponential decay) significantly outperforms all existing methods, achieving the best ADE of 0.34 and FDE of 0.47. On the UNIV dataset, the previous Social-STGCNN method by \cite{mohamed2020social} maintains the best performance with an ADE of 0.44 and FDE of 0.79. However, our method with Graph 2 (using field of view and $5m$ distance threshold for neighborhood definition) and Kernel 2 closely follows the best performance with an ADE of 0.45 and FDE of 0.82. Other graph variations and kernel functions also show very close performance on this dataset, further highlighting the importance of using geometric definitions. For the ZARA01 dataset, the Social-BiGAT method by \cite{kosaraju2019social} achieves the best ADE of 0.30, while our method with Graph 2 and Kernel 1 achieves the best FDE of 0.52. On the ZARA02 dataset, our method with Graph 2 and Kernel 2 outperforms all existing methods, achieving the best ADE of 0.27 and FDE of 0.45.

Overall, our method with Graph 1 and Kernel 1 achieves the best average ADE of 0.43 and FDE of 0.73 across all datasets, surpassing previous methods.
Our results suggest that methods incorporating attention mechanisms, whether through softmax (e.g., social-BiGAT by \cite{kosaraju2019social}) or geometric kernel functions, consistently enhance accuracy, underscoring the importance of kernel functions for graph representations in GNN methods. Furthermore, our results highlight the effectiveness of integrating domain knowledge, such as pedestrian's field of view, $5m$ distance threshold, and approach dynamics, into neighborhood definitions for modeling human interactions in crowded environments. While \cite{mohamed2020social} found no improvement with the exponential decay kernel, our results suggest that performance outcomes can vary and improve depending on the interaction neighborhood definition.

While no single method consistently outperforms others across all datasets, our proposed methods exhibit competitive performance. Performance discrepancies may arise from unaccounted factors like landmarks or scene information, highlighting the potential impact of external factors. Notably, \cite{lv2023ssagcn} demonstrated significant accuracy improvements by integrating scene information into their graph attention network.

Our primary objective was to demonstrate how incorporating domain knowledge can enhance data-driven trajectory prediction, rather than finding the best-performing model. Thus, we maintained the same parameters as the referenced work. Additionally, our model estimates uncertainty using the covariance matrix in the bivariate Gaussian distribution, capturing the inherent uncertainty in predictions. Future work can explore Bayesian approaches and the impact of data and domain uncertainties on model performance.

\subsection{Conclusion}
In summary, this case study introduces a geometric GNN architecture featuring various kernel functions and interaction topologies inspired by psychological studies. This framework lays a foundation for better understanding and modeling human interactions. Integrating a $180^ \circ$ field of view in the interaction neighborhood and employing the inverse of norm kernel function to weigh neighboring agents' importance contribute to overall accuracy improvements, reflected in enhanced ADE and FDE metrics across all datasets. Our results suggest that different topologies of interaction and kernel functions may perform better under different scenarios. However, determining the optimal designs that best explain complex interactions across domains remains an open challenge.

\bibliography{MECC_2024_Honarvar}             
                                                   







\end{document}